%% file: root.tex
\documentclass[letterpaper, 10 pt, conference]{ieeeconf}  

\IEEEoverridecommandlockouts                              

\usepackage{multicol}
\usepackage{multirow}
\usepackage[dvipsnames]{xcolor}
\usepackage{graphicx}
\usepackage{booktabs} 
\usepackage{soul,color} 
\usepackage{gensymb} 
\usepackage[norelsize, linesnumbered, ruled, lined, boxed, noend]{algorithm2e} 
\usepackage{comment}
\usepackage{amsmath}
\usepackage{diagbox}
\usepackage{hyperref}
\usepackage{adjustbox}
\SetArgSty{textup}

\SetCommentSty{mycommfont}
\usepackage{caption}
\usepackage{subcaption}
\SetAlgoSkip{}

\SetAlFnt{\small}
\SetAlCapFnt{\small}
\SetAlCapNameFnt{\small}
\SetAlCapHSkip{0pt}

\definecolor{mitred}{rgb}{0.64, 0.12, 0.20}
\definecolor{forestgreen}{rgb}{0.0, 0.27, 0.13}
\definecolor{auburn}{rgb}{0.43, 0.21, 0.1}

\title{\LARGE \bf
ASAP: Automated Sequence Planning for Complex Robotic Assembly with Physical Feasibility
}

\author{Yunsheng Tian$^{1,\dagger}$, Karl D.D. Willis$^{2}$, Bassel Al Omari$^{3,\dagger}$, Jieliang Luo$^{2}$, Pingchuan Ma$^{1}$, Yichen Li$^{1}$, \\ Farhad Javid$^{2}$, Edward Gu$^{1}$, Joshua Jacob$^{1}$, Shinjiro Sueda$^{4}$, Hui Li$^{2}$, Sachin Chitta$^{2}$ and Wojciech Matusik$^{1}$%
\thanks{$^{1}$MIT CSAIL
        \url{{yunsheng,pcma,yichenl,jmjacob,wojciech}@csail.mit.edu}, \url{egu@mit.edu}}%
\thanks{$^{2}$Autodesk Research
        \url{{karl.willis,rodger.luo,farhad.javid,hui.xylo.li,sachin.chitta}@autodesk.com}}%
\thanks{$^{3}$University of Waterloo
        \url{b2alomar@uwaterloo.ca}}%
\thanks{$^{4}$Texas A\&M University
        \url{sueda@tamu.edu}}%
    \thanks{$\dagger$Work partially done while interning at Autodesk Research.}%
}

\begin{document}

\maketitle
\thispagestyle{empty}
\pagestyle{empty}

\begin{abstract}
\input{0_abstract}
\end{abstract}

\input{1_introduction}
\input{2_related_work}
\input{3_sequence_planning}
\input{4_feasibility_check}
\input{5_evaluation}
\input{6_conclusion}

\section*{Acknowledgement}
We thank the MIT SuperCloud and Lincoln Laboratory for HPC resources, Jie Xu for discussing the initial idea, Yifei Li for help with experiments, Michael Foshey, Chao Liu and Branden Romero for help on the robotic setup. This work is funded by Autodesk, and in part by NSF CAREER-1846368.

\bibliographystyle{IEEEtran}
\bibliography{references}

\end{document}

%% file: 0_abstract.tex
The automated assembly of complex products requires a system that can automatically plan a physically feasible sequence of actions for assembling many parts together. In this paper, we present ASAP, a physics-based planning approach for automatically generating such a sequence for general-shaped assemblies. ASAP accounts for gravity to design a sequence where each sub-assembly is physically stable with a limited number of parts being held and a support surface. We apply efficient tree search algorithms to reduce the combinatorial complexity of determining such an assembly sequence. The search can be guided by either geometric heuristics or graph neural networks trained on data with simulation labels. Finally, we show the superior performance of ASAP at generating physically realistic assembly sequence plans on a large dataset of hundreds of complex product assemblies. We further demonstrate the applicability of ASAP on both simulation and real-world robotic setups. Project website: \textcolor{blue}{\href{http://asap.csail.mit.edu/}{asap.csail.mit.edu}}

%% file: 1_introduction.tex
\section{Introduction}

Real-world products often involve complex assemblies, necessitating specially designed assembly lines for their assembly, maintenance, and repair. In manufacturing, these {\em fixed} assembly lines are highly efficient for individual assemblies. However, creating them demands significant effort for design, programming, and setup, and they struggle to adapt to different assemblies. Particularly for high-mix, low-volume products, there is a need for automated assembly lines that are {\em flexible} and easily repurposed.

Assembly automation begins with understanding the sequence for assembling parts, typically done manually by experienced manufacturing teams. However, generating physically feasible assembly sequences remains a challenging research problem for several reasons:
(1) Sequence Planning: The number of potential assembly sequences grows exponentially with part count, with not all sequences being equally good or feasible. 
(2) Physical Feasibility: Planned sequences may not be physically executable due to unsatisfied physical constraints. For example, a collision-free path might not exist for a given assembly order, or parts could fall during assembly due to instability.
(3) Geometric Complexity: The geometry of objects can be highly complex, such as gears with numerous faces. Planning collision-free paths or assessing gravitational stability accurately in these scenarios is non-trivial.

\begin{figure}[t]
    \centering
    \includegraphics[width=0.48\textwidth]{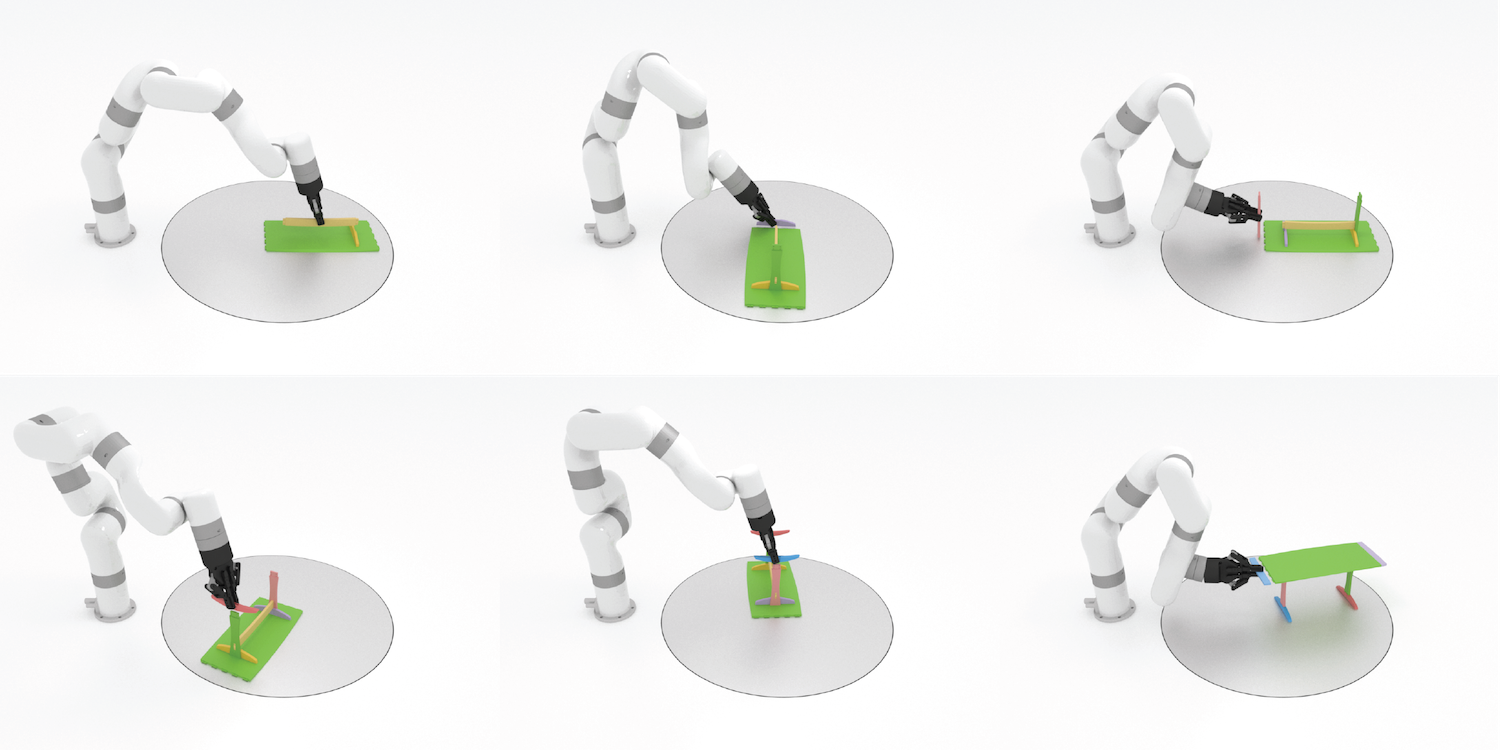}
    \caption{Assembly plans generated autonomously from ASAP for a desk positioned on a rotary table including physically feasible assembly sequences, collision-free paths, gravitationally stable poses, gripper grasps, and robot arm motion.}
    \label{fig:teaser}
\end{figure}

We tackle these challenges with \textbf{ASAP} (\textbf{A}utomated \textbf{S}equence Planning for Complex Robotic \textbf{A}ssembly with \textbf{P}hysical Feasibility).
To guarantee physical feasibility of assembly sequences, we use physics-based simulation to check for collision-free paths. We also introduce an efficient method to verify gravitational stability and identify parts to be held for stability. Additionally, we find the most stable pose when placing the assembly on a support surface.
To speed up planning and reduce the combinatorial search space, we employ the assembly-by-disassembly principle and efficient tree search algorithms. We guide the sequence search using geometric heuristics or graph neural networks (GNNs) trained on simulation-based data.

In summary, our work makes the following contributions:
\begin{enumerate}
    \item An efficient assembly sequence planning algorithm for generating physically feasible sequences for complex-shaped, contact-rich assemblies, ensuring both collision-free paths and gravitational stability.
    \item A learning-based approach for selecting the next part to disassemble, using GNNs trained on a large dataset of product assemblies.
    \item A greedy approximate gravitational stability checking algorithm that more efficiently determining unstable parts for stabilization compared to the combinatorially expensive naive approach.
    \item Evaluation on a large dataset of hundreds of complex product assemblies, demonstrating state-of-the-art performance compared to established baselines.
    \item Integration of grasp planning and robot arm inverse kinematics, guaranteeing feasible robotic execution in both simulation and real-world experiments.
\end{enumerate}

%% file: 2_related_work.tex
\section{Related Work}

\subsection{Assembly sequence planning}

To automatically determine feasible assembly sequences and paths, the non-directional blocking graph~\cite{halperin2000general} is proposed but limited to simple geometries due to computational complexity. Later, motion planning methods were introduced for complex-shaped objects either by geometric collision checking~\cite{sundaram2001disassembly, le2009path, zhang2020c} or leveraging physics simulation~\cite{tian2022assemble}. For real-world assemblies, gravitational stability is also considered during planning~\cite{abe1999stability, aleotti2009efficient}. However, evaluating the physical feasibility of complex-shaped assemblies accurately is difficult and time-consuming. To speed up the feasibility check, \cite{rodriguez2019iteratively} propose an iterative procedure on planar profile assembly by checking complex dynamic constraints later only if simpler static constraints are all satisfied.

To tackle the combinatorial complexity of sequence optimization, evolutionary methods have been proposed, such as genetic algorithms~\cite{kongar2006disassembly, marian2006genetic, tseng2018block}, ant colony optimization~\cite{wang2014mechanical}. However, these methods require evaluating massive sequences in order to be successful; hence their evaluation metrics are usually simple and do not consider gravitational stability, preventing the direct deployment of their optimized sequences in a real-world setup. More efficient search techniques have been explored in \cite{huang2016framefab, huang2021robotic} to specifically showcase stable assembly of frame shapes and bar structures.

In contrast to prior work, we demonstrate our sequence planner on a complex general-shaped 3D assembly dataset derived from the Fusion 360 Gallery Assembly Dataset~\cite{willis2022joinable} and a real mechanical assembly dataset~\cite{lupinetti2019content}, featuring more complex shapes than previously considered. While \cite{tian2022assemble} also use the dataset from \cite{willis2022joinable} to demonstrate a sequence planner, they only consider geometric motion constraints in a gravity-free setting and perform assembly in midair. 
In our work, we consider more physical constraints, such as gravity and friction, as well as robotic constraints, in determining the feasibility of each assembly step. Additionally, we take into account the use of a flat support surface, stable poses, and grippers for holding unstable parts.

\subsection{Physics-based simulation for assembly}

Unlike pure geometric analysis, physics-based simulation aids in assessing the physical feasibility of (dis)assembly sequences. Prior work, such as \cite{aleotti2009efficient} and \cite{rakshit2014influence}, employs physics engines for stability analysis in disassembly sequences. However, their experiments are limited to simple block-stacking scenarios instead of complex shapes. In \cite{kim2017parts}, physical simulation constructs a physically reasonable graph to speed up searches, tested on box-on-table and peg-in-hole tasks. Reinforcement learning with physical simulation is also used for learning robotic control policies~\cite{yu2021roboassembly, de2021autonomous, luo2021learning}, mainly focusing on simple insertion tasks like peg-in-hole or lap-joint due to limitations of convex-hull-based simulators. Recently, \cite{tian2022assemble} and \cite{narang2022factory} utilize simulators based on Signed Distance Field (SDF) to simulate contact-rich complex assemblies. However, none of these studies explore sequence generation for multi-part assemblies that are both stable and executable in the real world.

\subsection{Learning for assembly sequence planning}

Learning-based methods are promising to tackle the combinatorial problem of assembly sequence generation. Self-supervised learning \cite{zakka2020form2fit} collects data by trial-and-error kit disassembly but is limited to 2D shape-matching tasks. Reinforcement learning methods \cite{watanabe2020search, kitz2021neural, funk2022graph} can train effective 3D assembly planners but demonstrate on simplified physics or geometries due to high sample complexity. Our approach employs supervised learning inside planning, using a GNN to predict disassembly sequences with physical feasibility for complex 3D assemblies. We gather training data from diverse assemblies generated via physics-based simulation, and the network shows effective guidance for planning disassembly sequences on unseen assemblies.

%% file: 3_sequence_planning.tex
\section{Physically feasible sequence planning}
\label{sec:seq_planning}

Our objective is to plan an assembly sequence under physical constraints using individual parts and their assembled states as input. The output sequence details each assembly step, including the part to assemble, its motion path, assembly pose, and other parts to be held. Solving them jointly yields a complete and physically feasible sequence.

Searching in a combinatorial sequence space is challenging, especially with realistic physical constraints. To reduce the complexity, we employ the assembly-by-disassembly concept~\cite{demello1991}, obtaining assembly sequences from reverse disassembly sequences due to a bijection between them for rigid parts. In this section, we propose a tree search algorithm that efficiently prioritizes part disassembly by leveraging geometric heuristics or neural guidance, and identifies stable poses that satisfy physical constraints. 

\subsection{Disassembly tree search}
\label{sec:disassembly_tree_search}

\begin{figure}[t]
    \centering
    \includegraphics[width=0.30\textwidth]{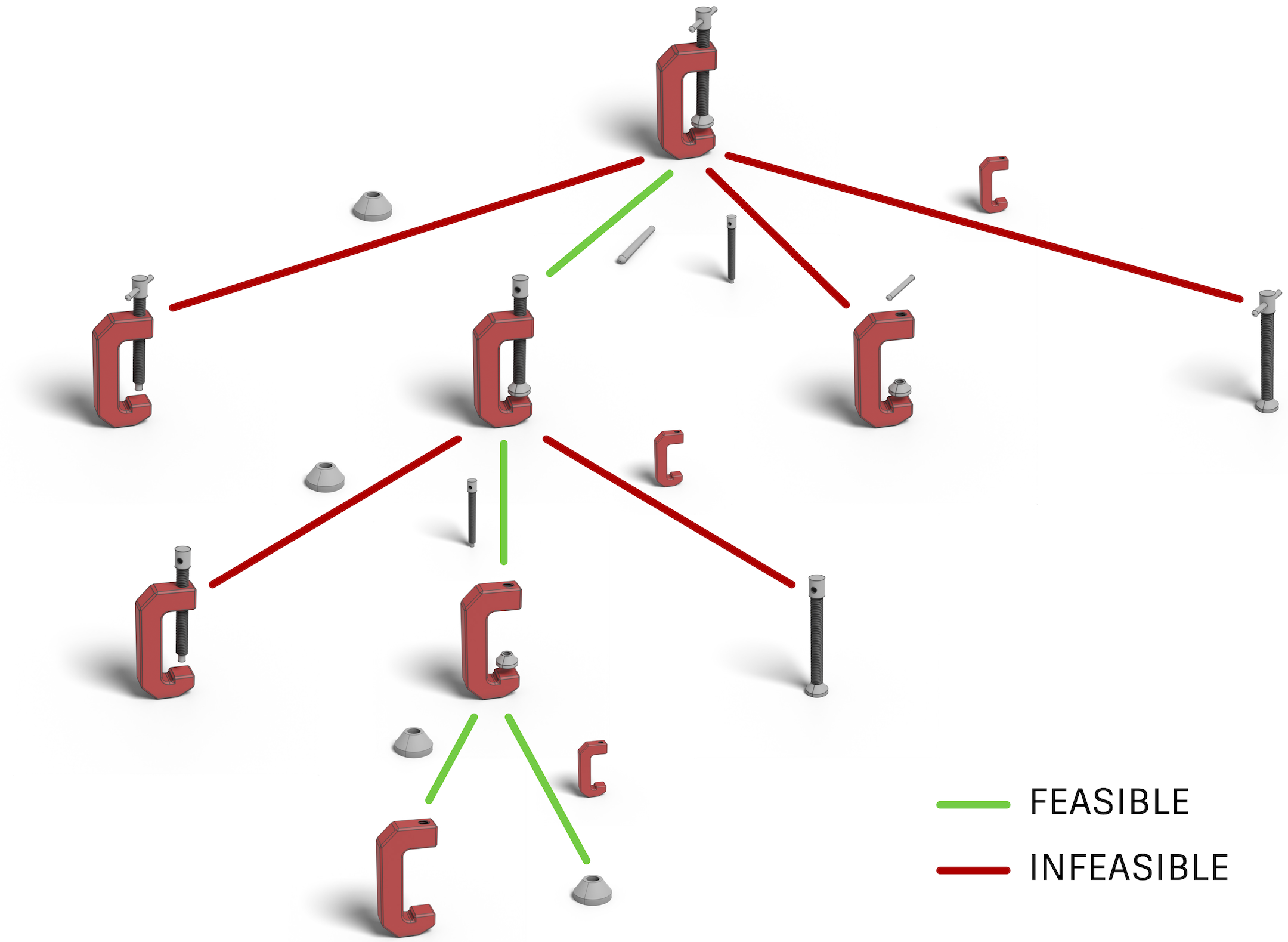}
    \caption{An example disassembly tree where nodes represent partial assemblies and edges represent feasible (green) and infeasible (red) disassembly actions.}
    \label{fig:disassembly_tree}
\end{figure}

We formulate the sequence planning as a tree-search framework, utilizing established techniques for efficient exploration. The core data structure is a disassembly tree, as shown in Fig.~\ref{fig:disassembly_tree}. Each path of the tree represents a particular way of how the assembly is disassembled from top to bottom.

\begin{algorithm}
    \caption{Physics-Based Disassembly Tree Search}
    \label{alg:tree_search}
    \SetAlgoLined
    \KwIn{Assembly $G_0 = \{p_1,...,p_n\}$, evaluation budget $N$.}
    \KwOut{A physically-feasible disassembly sequence.}
    $T \leftarrow$ EmptyTree(); $i = 0$;\\
    $T$.AddNode($G_0,$ valid=true);\\
    \While {$i < N$ and search not complete} {
        $G \leftarrow$ \textcolor{blue}{SelectNode$(T)$}; \tcp*[f]{\S\ref{sec:node_selection}}\\
        \For (\tcp*[f]{\S\ref{sec:part_selection}}) {part $p$ in \textcolor{blue}{SelectPart$(G)$}} {
            $G' \leftarrow G \setminus \{p\}$;\\
            \lIf{$T$.HasEdge($G, G'$)} {\textbf{continue}}
            \For (\tcp*[f]{\S\ref{sec:pose_selection}}) {pose $s$ in \textcolor{blue}{SelectPose$(G)$}} {
                $P_a \leftarrow$ \textcolor{mitred}{CheckAssemblable$(G', p, s)$}; \tcp*[f]{\S\ref{sec:check_path}}\\
                $P_g \leftarrow$ \textcolor{mitred}{CheckStable$(G', s)$}; \tcp*[f]{\S\ref{sec:check_stability}}\\
                $i = i + 1$;\\
                \If {$P_a \neq$ null and $P_g \neq$ null} {
                    $T$.AddNode($G',$ valid=true);\\
                    $e \leftarrow T$.AddEdge($G, G',$ valid=true);\\
                    $e$.StoreInfo($s, P_a, P_g$);\\
                    \textbf{break}; \tcp*[f]{feasible step}
                }
            }
            \uIf {not $T$.HasNode($G'$)} {
                $T$.AddNode($G',$ valid=false);
            }
            \uIf {not $T$.HasEdge($G, G'$)} {
                $T$.AddEdge($G, G',$ valid=false);
            } 
            \uElseIf (\tcp*[f]{complete sequence}) {$|G'| = 1$} {
                \Return{$T$.FindPath($G_0, G'$)};
            }
            \textbf{break};
        }
    }
    \Return{null};
\end{algorithm}

Algorithm~\ref{alg:tree_search} outlines our disassembly tree search framework. Given the fully assembled assembly $G_0$ with $n$ parts, our goal is to find a physically feasible disassembly sequence within the total simulation evaluation budget $N$. 
In our algorithm, we build and iteratively update a disassembly tree until a feasible sequence is found. The success of the algorithm relies on the following implementations to expand the tree in each iteration, as highlighted in Algorithm~\ref{alg:tree_search}:
\begin{itemize}
    \item \textcolor{blue}{SelectNode($T$)}: which node to expand tree $T$ from (i.e., which partial assembly to disassemble). 
    \item \textcolor{blue}{SelectPart($G$)}: which part to disassemble from node $G$.
    \item \textcolor{blue}{SelectPose($G$)}: under which pose to disassemble $G$.
\end{itemize}
In each iteration, we first select a node $G$ from the tree $T$ to expand, guided by the specific tree search method in SelectNode($T$). After that, we decide which part $p$ to disassemble from $G$ using SelectPart($G$). The remaining assembly without $p$ is referred to as $G'$. Finally, we select a pose $s$ to orient $G$ and attempt to disassemble part $p$.

The feasibility of disassembly is determined by: (1) \textcolor{mitred}{CheckAssemblable()}, which gives a disassembly motion plan $P_a$ (\S\ref{sec:check_path}) and (2) \textcolor{mitred}{CheckStable()}, which gives a part holding plan $P_g$ to ensure stability (\S\ref{sec:check_stability}). If both procedures return feasible solutions, the tree expansion is successful. Otherwise, we continue searching for new poses and different paths to expand the tree until either a feasible sequence is found, or the budget is reached, or the tree is fully expanded.

\subsection{Node selection}
\label{sec:node_selection}

Any standard tree traversal method can be applied to select a node to expand in each iteration, i.e., SelectNode() in Algorithm~\ref{alg:tree_search}. For simplicity, we apply a standard implementation of depth-first search (DFS) that explores the tree as far as possible along each branch before backtracking.

\subsection{Part selection}
\label{sec:part_selection}

\subsubsection{Geometric heuristic}
To wisely select a part to disassemble, we leverage the insight that exterior parts are generally easier to disassemble due to fewer precedence constraints. We use the distance between a part's center of mass and the assembly center (bounding box center) to prioritize outside-in part selection.

\subsubsection{Learning-based}
\label{sec:part_selection_learning}
We also implement a learning-based approach to accelerate the tree search described in Algorithm~\ref{alg:tree_search}. Our network is trained by supervision on the task of node classification to predict the next part to disassemble. 6,368 disassembly sequence labels are generated from 1,906 training assemblies out of 2,146 total assemblies from the dataset described in \S\ref{sec:eval_data} using simulation following a dynamic beam search to balance data diversity and computational efficiency. Inspired by previous works on learning for furniture assembly~\cite{aslan2022assemblerl,li2020impartass} and CAD modeling~\cite{willis2022joinable}, we implement a GNN model that takes as input a graph $G$ with $n$ parts $\{p_1, p_2, ..., p_n\}$ forming graph nodes connected by adjacency. The network outputs a probability distribution over all nodes, indicating the likelihood of each part being next in the disassembly sequence.

\begin{figure}[t]
    \centering
    \includegraphics[width=0.40\textwidth]{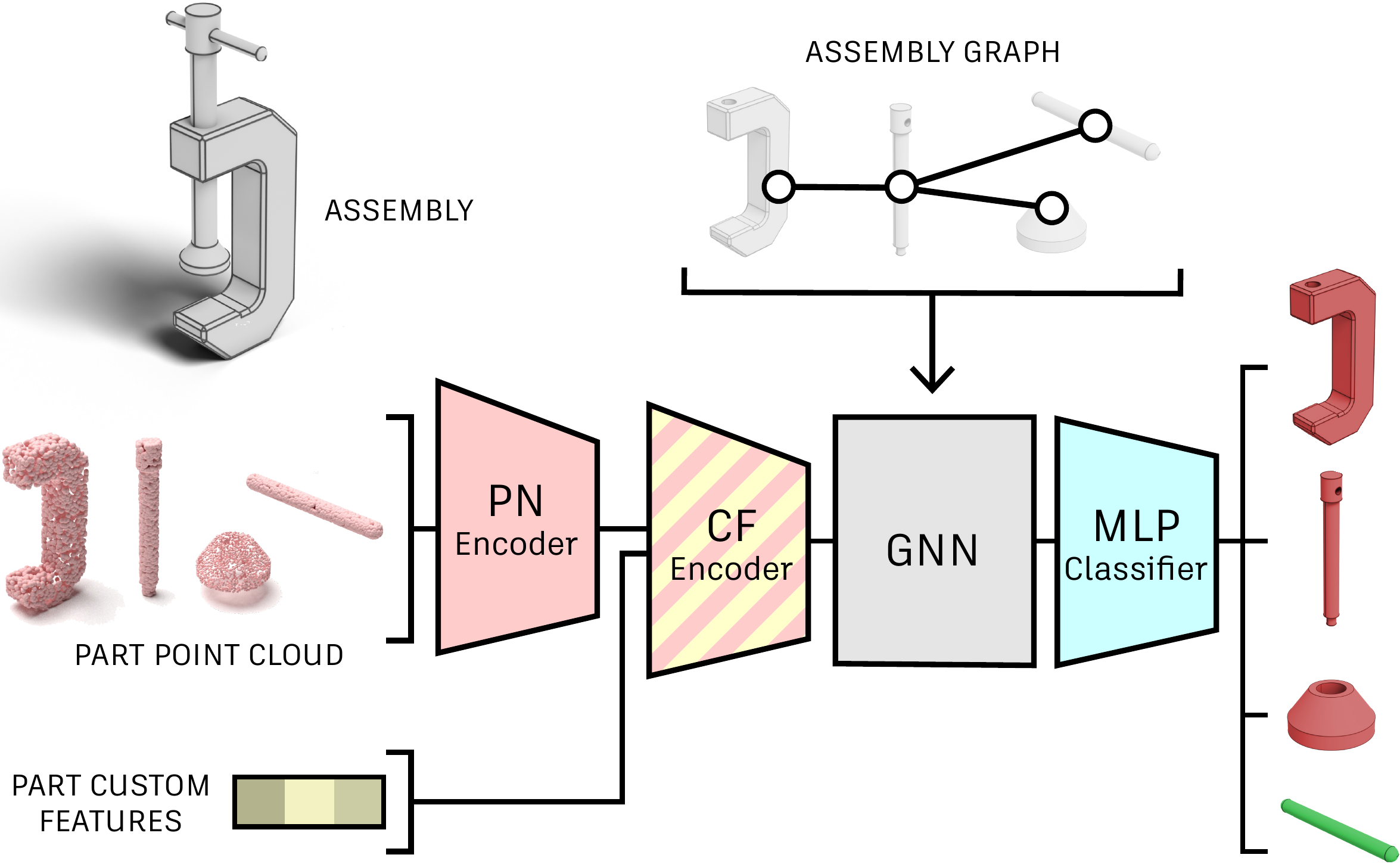}
    \caption{Network architecture for learning part disassembly priority. Given an assembly as input, a graph neural network is used to predict the next part to be removed.}
    \label{fig:network}
\end{figure}

Fig.~\ref{fig:network} illustrates the network architecture consisting of four modules: a PointNet encoder (PN encoder)~\cite{qi2017pointnet}, a custom feature encoder (CF encoder), GNN message passing layers, and an MLP classifier. The two encoders extract two kinds of per-part features. The PN encoder extracts part geometric features from uniformly sampled point clouds (1K points per part), and the CF encoder provides additional information, such as part volume, part distance to the assembly center, and the number of adjacent parts. The two kinds of features together form a graph node embedding. Two rounds of message passing are used to update the graph node features between connected neighboring parts. Finally, the MLP classifier predicts the next part to disassemble from the final graph node embedding. We use a standard binary cross-entropy loss as the supervision signal between the predicted probability to disassemble a given part next $\hat{p}_i$ and the ground truth label $p^{gt}_i$ indicating the next part to disassemble.

\subsection{Pose selection}
\label{sec:pose_selection}

Since it is impractical to evaluate infinite poses in a 3D space for a given assembly, we propose to use quasistatic stable poses~\cite{goldberg1999part} of assembly $G$ as the pose candidates to select from (top-5 is empirically good enough), which provides a much reduced and reasonable search space. Furthermore, we observe that reusing poses during the assembly generates more intuitive sequences and also improves success rate.

%% file: 4_feasibility_check.tex
\section{Feasibility check for assembly sequences}
\label{sec:feasibility_check}

This section describes how we leverage physics-based simulation to check whether an assembly sequence is physically feasible. First, \S\ref{sec:simulation} describes our physics-based assembly simulation; Then, \S\ref{sec:check_path} focuses on finding collision-free paths for parts to be (dis)assembled using a physics-based path planning approach. Finally, \S\ref{sec:check_stability} describes our method to check whether a given (partial) assembly can be gravitationally and frictionally stable, taking into account the support surface and a subset of parts to be held.

\subsection{Physics-based simulation for assembly}
\label{sec:simulation}

Our physics-based model is designed to work robustly with complex geometry for assembly and is able to handle minor problems in the input geometry, including small overlaps, potentially conflicting collision normals, and slight gaps.
We use the penalty-based contact model for the contact normal force and the frictional forces \cite{geilinger2020add,xu2021end}. We use implicit Euler to integrate the system forward, which allows us to handle these penalty forces robustly and stably. The penetration distance is computed accurately and efficiently with pre-computed signed distance fields (SDF).

\subsection{Disassembly path planning}
\label{sec:check_path}

To check whether a part can be assembled, we leverage again the idea of assembly-by-disassembly where disassembly path planning can be applied to connect the part's assembled state to a disassembled state. Motivated by the recent success of physics-based path planning, we adopt \cite{tian2022assemble} to generate a collision-free assembly path for each part. Specifically, the planner starts from the assembled state and searches for a sequence of forces to apply until a disassembled state or a predefined budget is reached.

\subsection{Gravitational stability check}
\label{sec:check_stability}

Existing works using physics simulation to check stability are limited to simple block-stacking scenarios with a fixed pose on the support surface~\cite{aleotti2009efficient, rakshit2014influence}. However, complex real-world assemblies usually require at least two hands for assembly and re-orientation since the stable pose might change as more parts are assembled. To address this challenge, we develop an efficient physics-based stability check algorithm for a given assembly $G$ with $n$ parts under a specific pose $s$, represented as a homogeneous transformation matrix. The algorithm outputs the set of maximum $M$ parts to be held to make $G$ stable under pose $s$ or returns null if it needs to hold more than $M$ parts to stabilize $G$. 

\begin{algorithm}[t]
    \caption{Physics-Based Stability Check}
    \label{alg:stability_check_greedy}
    \SetAlgoLined
    \KwIn{Assembly $G = \{p_1,p_2,...,p_n\}$ with pose $s$, max $M$ parts to be held, max simulation steps $N$, stable moving distance threshold $d_{th}$.}
    \KwOut{Parts to be held to make $G$ stable with pose $s$.}
    $P_g \leftarrow \{\}$;\\
    \While {$|P_g| \leq M$} {
        ResetSimulation($G, s, P_g$);\\
        \lFor {$i$ in $1,...,n$} {$\textbf{q}_{i_0} \leftarrow$ GetPosition($p_i$)}
        $stable \leftarrow$ true;\\
        \For (\tcp*[f]{$N$-step stability check}) {$j$ in $1,...,N$} {
            Simulate($\Delta t$); \tcp*[f]{run physics simulation}\\
            \For {$i$ in $1,...,n$} {
                $\textbf{q}_{i_j} \leftarrow$ GetPosition($p_i$);\\
                \If {$\|\textbf{q}_{i_j} - \textbf{q}_{i_0}\| > d_{th}$ or IsDisconnected($p_i, G$)} {
                    $stable \leftarrow$ false; \tcp*[f]{$p_i$ falls}\\
                    $P_g \leftarrow P_g \cup \{p_i\}$; \tcp*[f]{hold $p_i$}\\
                    \textbf{break};
                }
            }
            \lIf {not $stable$} {\textbf{break}}
        }
        \lIf (\tcp*[f]{stable for $N$ steps}) {$stable$} {\Return{$P_g$}}
    }
    \Return{null}; \tcp*[f]{unstable}
\end{algorithm}

A naive way to discover feasible part-holding plans is by trying all $O(n^M)$ combinations of parts to hold, which is impractical for large assemblies. Therefore, we introduce a greedy approximation with $O(M)$ part-holding plans to search in Algorithm~\ref{alg:stability_check_greedy}. The key idea is to start with no parts held and only hold parts when we observe a part falling until $M$ parts are held. We use $P_g$ to denote the set of parts we hold to ensure stability, which is an empty set initially and gets iteratively updated. In each iteration, we obtain the initial position of all the parts, denoted as $\textbf{q}_{i_0}$ for part $p_i$. Next, we run the physics simulation for maximum $N$ time steps to check stability. In each step $j$, we obtain the updated position $\textbf{q}_{i_j}$, and calculate the Euclidean distance between $\textbf{q}_{i_j}$ and initial position $\textbf{q}_{i_0}$. If the distance is larger than a threshold $d_{th}$, this means part $p_i$ is falling and thus not stable. Apart from that, we also check if part $p_i$ loses contact with its neighboring parts; if so, it is also unstable. If any part is unstable in step $j$, we add part $p_i$ to $P_g$ and restart the whole process. The algorithm stops until either a feasible $P_g$ is found, or we cannot hold extra parts ($|P_g| > M$). Note that we use position change as the stability check criterion because it is smoother in time for contact-rich simulation, as opposed to numerically noisy velocity or acceleration in previous works~\cite{aleotti2009efficient, rakshit2014influence}.

%% file: 5_evaluation.tex
\section{Evaluation}

\subsection{Dataset for assembly sequence planning}
\label{sec:eval_data}

\begin{table*}[t]%
\small
\caption{Success rate (\%) comparison of ASAP on the assembly test dataset against several baseline methods.}
\vspace{-3mm}
\label{tab:success_rate}
\begin{center}
\begin{adjustbox}{width={\textwidth}}
\begin{tabular}{cc|ccc|ccc}

  \toprule
  \multicolumn{2}{c|}{\multirow{2}{*}{\textbf{Method}}} & \multicolumn{3}{c|}{\textbf{Success Rate (\%) (Low Budget)}} & \multicolumn{3}{c}{\textbf{Success Rate (\%) (High Budget)}}\\
   & & 2 Parts Held & 3 Parts Held & 4 Parts Held & 2 Parts Held & 3 Parts Held & 4 Parts Held \\
  \midrule
  \multirow{2}{*}{\textbf{ASAP (Ours)}} 
  & Heuristics & 51.25 & 61.25 & 68.75 & 66.67 & 74.17 & 80.83 \\
  & Learning & \textbf{54.58} & \textbf{62.92} & \textbf{69.58} & \textbf{67.08} & \textbf{76.25} & \textbf{82.08} \\
   \midrule
  \multirow{3}{*}{\textbf{Baseline}} & Random Permutation & 14.58 & 25.42 & 41.25   & 27.92 & 43.33 & 55.42 \\
  & Genetic Algorithm~\cite{kongar2006disassembly} & 14.17 & 25.83 & 40.00  & 30.83 & 41.25 & 51.25 \\
  & Assemble Them All~\cite{tian2022assemble} & 19.17 & 27.08 & 35.42   & 30.42 & 46.25 & 56.67 \\
  \bottomrule

\end{tabular}
\end{adjustbox}
\vspace{-7mm}
\end{center}
\end{table*}%

We build a large complex 3D assembly dataset derived from the Fusion 360 Gallery Assembly Dataset~\cite{willis2022joinable} and a real mechanical assembly dataset~\cite{lupinetti2019content}, consisting of 2,146 assemblies from 3 to 50+ parts with more complex shapes than those used previously for this problem. To ensure feasibility, we perform data filtering processes to make sure part meshes are watertight and able to be (dis)assembled without physical deformation, and remove parts that intersect with others in the assembly greater than a given distance. Among these assemblies, we select 240 assemblies to be the test set for benchmarking using the following procedures: First, we filter out all potentially unstable assemblies in the dataset by loading them into simulation and checking for stability using the top stable poses generated by \S\ref{sec:pose_selection}. Next, we group the remaining dataset by the number of parts within the assembly and randomly sample up to 10 assemblies from each group to ensure a suitable representation of complex assemblies in the distribution. Finally, we combine all sampled assemblies to form the test set and the remaining 1,906 assemblies become our training set.

\subsection{Baseline methods}
\label{sec:baseline}

We compare ASAP with a na\"ive baseline and two established methods that generate and optimize (dis)assembly sequences: (1) \textit{Random Permutation} generates completely random (dis)assembly sequences and evaluates them until the budget is met, to serve as a lower bound on performance. (2) \textit{Genetic Algorithm}~\cite{kongar2006disassembly, tseng2018block} starts from randomly permuted sequences and iteratively optimizes the sequences based on a {\it fitness function}, defined as the number of steps of the sequence that are physically feasible. (3) \textit{Assemble Them All}~\cite{tian2022assemble} similarly leverages physics simulation to find assembly sequences by randomly selecting parts to disassemble and checking for collision-free disassembly paths. However it does not consider stability as a criterion, but assumes a gravity-free setting with no supporting surface and parts to be held. For comparison, we run \textit{Assemble Them All} repeatedly to generate sequences until the budget is met. To take into account physical feasibility, we use our stability check algorithm in \S\ref{sec:check_stability} under the top poses searched by the quasistatic pose estimator in \S\ref{sec:pose_selection} to evaluate the feasibility of the generated sequences for all baseline methods.

\subsection{Experimental setup}

We benchmark two variants of our ASAP algorithm with different part selection methods (see \S\ref{sec:part_selection}). We run our experiments in parallel on Intel Xeon Platinum 8260 CPUs with 4GB RAM per core, where each experiment for a single assembly runs on a single CPU core. We set a 2-hour timeout budget for each experiment. 
For network implementation, all modules use 512 hidden neurons, ReLU activation, and batch normalization. The GNN uses two GATv2~\cite{brody2021attentive} message-passing layers with eight heads. We withhold the test set described in \S~\ref{sec:eval_data} and split the remaining data 90:10 for training and validation. Training takes 5 hours using an NVIDIA V100 GPU for 150 epochs, with a batch size of 160 and a learning rate of $0.001$. We use the Adam optimizer with a scheduler that reduces the learning rate on the plateau.

\subsection{Quantitative success rate comparison}
\label{sec:quantitative}

We now compare the performance of our algorithms against the baseline methods using the \textit{success rate} metric: the percentage of assemblies from the test dataset that can be successfully (dis)assembled, given a fixed simulation evaluation budget and a set number of parts to be held. To represent a realistic scenario, we show results for both a low budget (50 evaluations) and a high budget (400 evaluations) in Table~\ref{tab:success_rate}, holding 2, 3, and 4 parts respectively.

The results show that both variants of our algorithm outperform the three baseline methods by a significant margin, thanks to the tree search formulation and efficient guidance on part selection that greatly reduces the search space. We observe improvements for all methods if allowing more parts to be held or given more evaluation budget, though a setup with two part holders and a limited budget is more practical. The outstanding performance of our \textit{Learning} variant shows that the neural guidance trained on a diverse set of assemblies with simulation-generated labels generalizes well to unseen and more challenging assemblies in the test set.

We report the median runtime of ASAP with a single CPU thread on different sizes of assemblies:
$\leq$5 parts: 43s;
(5,10] parts: 97s;
(10, 20] parts: 431s;
(21, 30] parts: 1705s;
$>$30 parts: 4190s.
Significant speed up can be achieved through parallelization in several components of Algorithm~\ref{alg:tree_search}.

We conduct quantitative comparisons without constraints from robotic manipulators, which isolates the impact of planning strategies on assembly success while abstracting away the nuances of robotic hardware and specialized tools. For methods and results on assembly planning with ASAP using a parallel gripper and a robot arm, please see \S\ref{sec:robotic_demo}.

\subsection{Qualitative assembly sequence comparison}
\label{sec:qualitative}

\begin{figure}[t]
    \centering
    \includegraphics[width=0.48\textwidth]{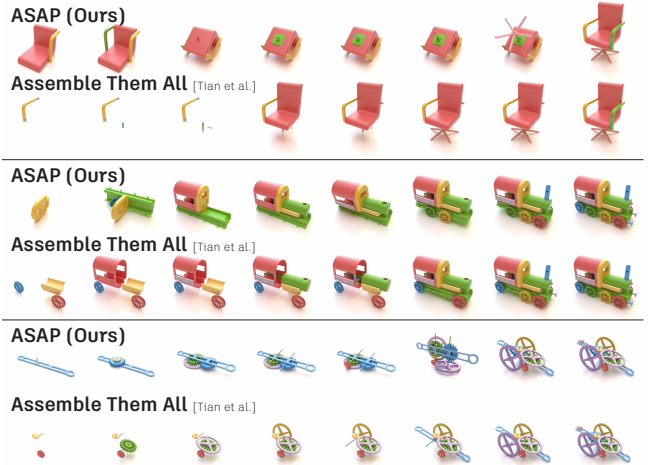}
    \caption{Assembly sequence comparison between ASAP and \textit{Assemble Them All}~\cite{tian2022assemble} on camera, train, and gear set. 
    }
    \label{fig:seq_comp}
\end{figure}

Fig.~\ref{fig:seq_comp} shows a comparison between assembly sequences generated by our method and \textit{Assemble Them All}. As described in \S\ref{sec:baseline}, \textit{Assemble Them All} does not take gravitational stability into consideration in assembly sequence generation, thus producing sequences with many floating parts and apparently unstable poses. In contrast, ASAP generates physically feasible sequences with a stability guarantee, which is more applicable to a real-world assembly setup.

\subsection{Ablation studies}

\begin{table}[t]
\small
\caption{Accuracy and speed up of the proposed stability check algorithm compared to the combinatorial ground truth (\textit{TP} = true positive, \textit{TN} = true negative, \textit{FN} = false negative).
}
\vspace{-2mm}
\label{tab:greedy_speedup}
\begin{center}
\begin{tabular}{c|cccc|c|c}
  \toprule
  {\# Parts} & TP & TN & FN & Timeout & Acc. & \multirow{2}{*}{Speed Up} \\
  {to Hold} & (\%) & (\%) & (\%) & (\%) & (\%) & \\
  \midrule
  2 & 58.5 & 19.2 & 9.6 & 12.7 & 89.0 & 13.90x\\
  3 & 66.3 & 8.4 & 7.8 & 17.5 & 90.5 & 17.03x\\
  4 & 69.8 & 4.0 & 4.1 & 22.1 & 94.7 & 23.04x\\
  \bottomrule
\end{tabular}
\vspace{-2mm}
\end{center}
\end{table}

We compare our greedy stability check algorithm (Algorithm~\ref{alg:stability_check_greedy}) to the full combinatorial check as the ground truth. We perform the comparison using 1,000 pairs of partial assemblies and poses randomly sampled from the data collected during simulation label generation. Table~\ref{tab:greedy_speedup} shows the accuracy of our greedy approximation.
Due to the complexity of the combinatorial algorithm, we limit each comparison with a 1-hour timeout and report the percentage of timeouts. The greedy algorithm achieves around 90\% accuracy and an order of magnitude speed-up, which becomes more significant as the number of parts that can be held increases due to a more expensive combinatorial check.

We also implemented a custom beam search for node selection with dynamically growing beam width, which performs similarly as DFS. Regarding geometric heuristics for part selection, we found the outside-in selection scheme introduced in \S\ref{sec:part_selection} offers a slight advantage over prioritizing parts with small volume or few adjacent parts.

\subsection{Robotic demonstration}
\label{sec:robotic_demo}

We integrate ASAP with a robotic setup for real-world deployment. The motion of the robotic arm and gripper is governed by grasp planning, inverse kinematics (IK) and collision detection. To determine feasible grasping points on an object using a parallel gripper, we sample 100 pairs of antipodal points on its surface through ray casting and surface normal comparisons to ensure force closure. Each pair of antipodal points guides the placement of the gripper’s fingertips, along with the selection of 5 orientations that positively align with the assembly direction. Next, we execute the assembly path with the chosen grasping pose. At each step of the path, we calculate the robot arm's joint angles via IK, using the gripper pose as input. A rotary table can be used to improve arm reachability, with the rotation angle optimized based on the desired gripper location and orientation. This process also includes thorough collision checking to ensure the robot arm, gripper, assembly parts, and the support surface remain collision-free. We prioritize grasp planning with poses that enhance stability and iterate until a collision-free and IK-feasible assembly plan is achieved. 

Following the above approach, ASAP generates feasible robotic assembly plans for diverse assemblies (e.g., Fig.~\ref{fig:teaser}). While comprehensive simulated results can be found on our \href{http://asap.csail.mit.edu/}{project website}, here we demonstrate the sim-to-real transfer on a real-world hardware setup with a 3D printed beam assembly with 5 parts. The central component of the hardware setup is a UFACTORY xArm 7 robotic arm, which utilizes a Robotiq 2F-140 gripper. To aid in the spatial localization of assembly components, we use a laser-cut placemat that allows the robot to determine the precise positioning of parts, thereby reducing potential assembly errors.

\begin{figure}[t]
    \centering
    \includegraphics[width=0.48\textwidth]{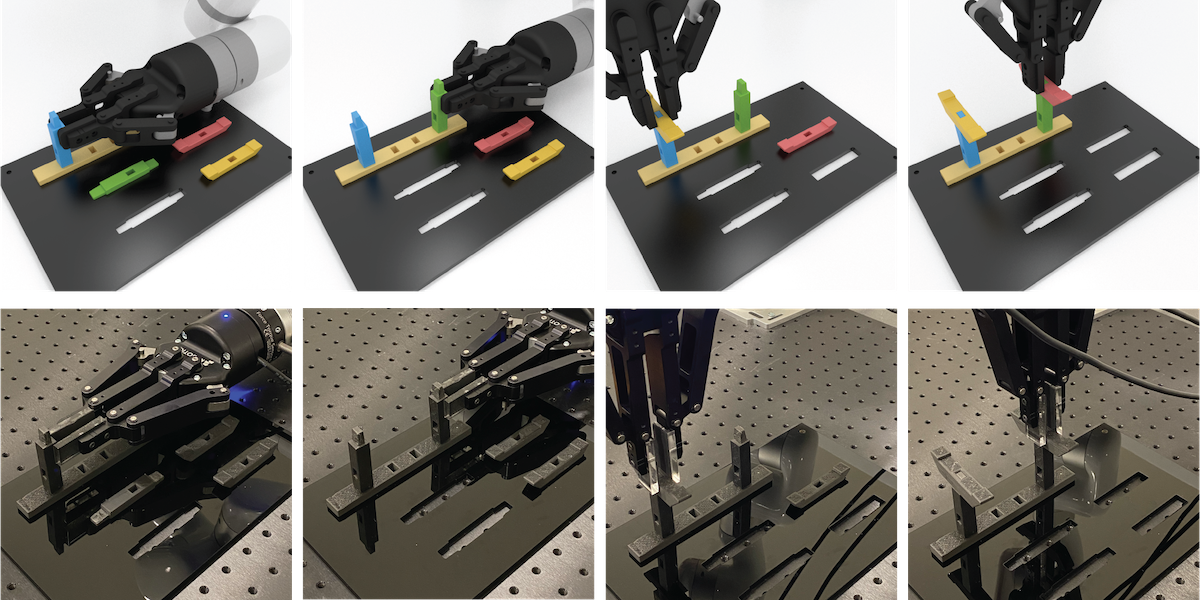}
    \caption{Robotic assembly plans of 3D printed beams generated by ASAP executed in simulation (top row) and real-world (bottom row).}
    \label{fig:real_demo}
\end{figure}

The direct sim-to-real transfer is non-trivial because due to tight millimeter-level clearances in assembly joints for stability and inherent errors in part fabrication and arm localization. After extensive calibrations, in Fig.~\ref{fig:real_demo}, we show a successful step-by-step correspondence between simulated ASAP plans and real hardware execution. The sim-to-real transfer can be made more robust by incorporating vision or force feedback and adaptive manipulation skills~\cite{chhatpar2001search, triyonoputro2019quickly}.

%% file: 6_conclusion.tex
\section{Limitations and Future Work}

We introduced ASAP, an efficient sequence planning algorithm for automated assembly that generates physically-feasible assembly plans for complex-shaped assemblies. We observe three notable failure cases of ASAP: (1) The disassembly tree is completely explored, but no feasible sequence is found, e.g., loosely assembled parts where stability is hard to guarantee or need extra grippers for unstable parts. (2) Physics evaluation can be slow when the contact dynamics in assemblies is complex. (3) Assemblies having strict precedence constraints may only have a few feasible sequences. In this case, ASAP fails to find a solution under a limited budget. The ratios of these failure cases are 50\%/45\%/5\%.

In future work, we plan to explore generating and executing more diverse assembly sequences in the real world using a more flexible robotic setup. To this end, we are interested to integrate different manipulation tools such as suction grippers and screwdrivers, and explore imitation learning or reinforcement learning to train physical assembly skills that are adaptable to real-life assembly tasks with uncertainties.